\documentclass[letterpaper, 10 pt, conference]{ieeeconf}  

\IEEEoverridecommandlockouts                              

\usepackage{graphics} 
\usepackage{epsfig} 
\usepackage{mathptmx} 
\usepackage{times} 
\usepackage{amsmath} 
\usepackage{amssymb}  
\usepackage{multirow}
\usepackage{color,soul}
\usepackage{algorithm}
\usepackage{siunitx}
\usepackage{algpseudocode}

\usepackage{hyperref}
\usepackage[comma,numbers]{natbib}

\title{\LARGE \bf
A Fast, Autonomous, Bipedal Walking Behavior over Rapid Regions
}

\author{Duncan Calvert$^{1,2}$, 
Bhavyansh Mishra$^{2}$, 
Stephen McCrory$^{1,2}$, 
Sylvain Bertrand$^{1}$,
Robert Griffin$^{1,2}$ \\
and Jerry Pratt$^{1,2}$
\thanks{This work was funded through ONR Grant N00014-19-1-2023, NASA Grant No. 80NSSC20M0197, and ARL Cooperative Agreement W911NF-21-2-0241.}
\thanks{$^{1}$The authors are with the Florida Institute for Human and Machine Cognition, 40 S Alcaniz St, Pensacola, FL 32502, United States}%
\thanks{$^{2}$Author is with the University of West Florida, 11000 University Pkwy, Pensacola, FL 32514, United States}%
\thanks{Email : \url{{dcalvert, bmishra, smccrory, sbertrand, rgriffin, jpratt}@ihmc.org}
}} 

\begin{document}

\maketitle
\thispagestyle{empty}
\pagestyle{empty}

\begin{abstract}
In trying to build humanoid robots that perform useful tasks in a world built for humans, we address the problem of autonomous locomotion. Humanoid robot planning and control algorithms for walking over rough terrain are becoming increasingly capable. At the same time, commercially available depth cameras have been getting more accurate and GPU computing has become a primary tool in AI research. In this paper, we present a newly constructed behavior control system for achieving fast, autonomous, bipedal walking, without pauses or deliberation. We achieve this using a recently published rapid planar regions perception algorithm, a height map based body path planner, an A* footstep planner, and a momentum-based walking controller. We put these elements together to form a behavior control system supported by modern software development practices and simulation tools.
\end{abstract}

\section{Introduction}
\label{introduction}

The humanoid form has tremendous potential in a world designed for humans. However, humanoid robots have struggled to continuously move and react to changing terrain when trying to reach a goal. They often pause for long periods of time while walking and are typically unresponsive to even large changes in terrain while walking. To advance humanoid capabilities so that they can respond to these changes and move without pausing, we present a behavior system that incorporates active perception and planning while walking. The presented version of this behavior works well for forward traversal on rough terrain, but perception field of view limits sideways and backward movement and turning in place.

While bipedal locomotion on flat and gently sloping terrain has entered a state of relative maturity\citep{reher2016realizing}\citep{Hobart_2020}\citep{Gibson_2021}, we and others have been working on increasing the capabilities over rough terrain\citep{Stumpf_2021}\citep{Fallon_2015}\citep{Griffin_2019}. In this paper, we focus on terrain that has spatial and angular discontinuities but still has good footholds, such as cinder block piles and broken concrete. This category of terrain lends itself to solutions that utilize planar region extraction and careful footstep planning.

To tackle the locomotion problem over this class of terrain, a stack of reliable technology is required. The robot hardware must have human-like range of motion, especially in the pitching of the ankles. It must be strong enough to perform lunges over multi-level and angled cinder block configurations. To withstand falls and long testing and development sessions, it must be durable and reliable. For software, we practice test-driven development to address reliability and cost of maintenance as lines of code grow\citep{Beck_2002}\citep{Ogheneovo_2014}. We use the Coactive Design method so that behavior operation is observable, predictable, and directable\citep{Johnson_2014}.  Finally, we use an advanced simulation environment to develop and test the coordination of control, communication, and user interface. Our simulation environment includes realtime perception which enables test-driven development of the entire system when the real robot is not available. This has enabled us to develop improvements and add features offline without introducing regressions.

\begin{figure}[!t]
\centering
    \includegraphics[width=1.0\columnwidth]{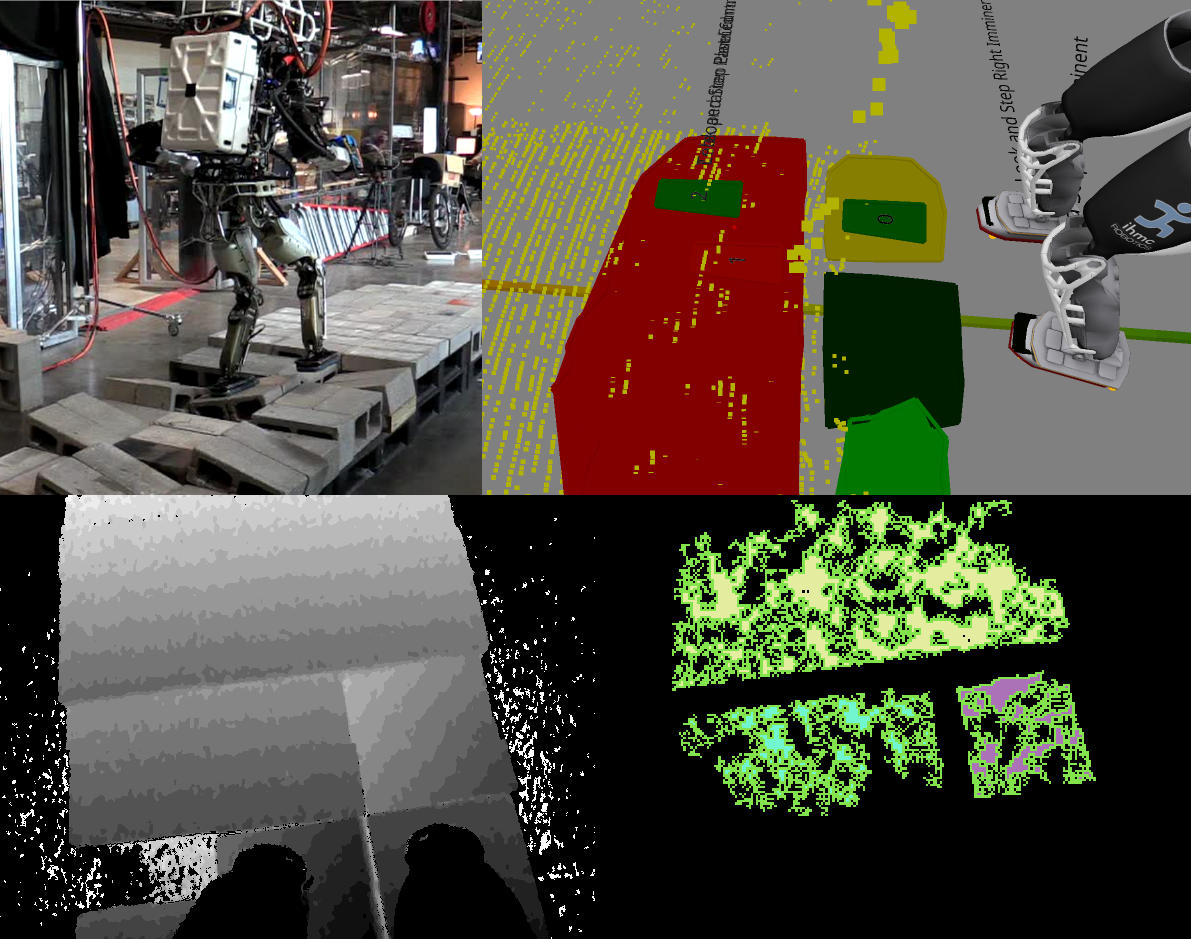}
\caption{The look-and-step behavior in action. A DRC Atlas autonomously walks to a goal at the other side of rough terrain. In the top right, the next three steps are shown planned. The depth camera image is shown in the bottom left and the planar region segmentation intermediary image shown in the bottom right.}
\vspace{-6mm}
\label{fig:look_and_step_highlight_composite}
\end{figure}

In this paper, we build on a realtime planar region perception algorithm\citep{Mishra_2021}, a height map based body path planner\citep{McCrory_2022}, a humanoid footstep planner for rough terrain\citep{Griffin_2019}, and a momentum-based control framework\citep{Koolen_2016} to present the contributions of this paper:
\begin{enumerate}
    \item A behavior architecture that coordinates perception, body path planning, footstep planning, and walking in realtime over rough terrain.
    \item A discussion of the challenges associated with realtime planning and walking with active perception.
    \item An operator interface that employs Coactive Design to offer shared autonomy and accelerate development.
    \item A description of the simulation environment used to develop this behavior.
\end{enumerate}

\section{Related Work}

Autonomous and reactive locomotion over various types of terrain has been studied since humanoid robots have existed. However, when falling is not an option, such as in the DARPA Robotics Challenge, these continuously re-planning systems were not used\citep{Stumpf_2014}\citep{Atkeson_2018}\citep{Johnson_2017}. Instead, success relied on human operator placed or adjusted footsteps, slowing the robots down and requiring careful supervision. We still strive to make bipedal locomotion fast, autonomous, reliable, and capable of traversing all types of terrain. We cite notable efforts that address this issue directly and that make improvements to the perception, planning, and control elements required to build this functionality.

In 2003, Kuffner et al. presented one of the first algorithms for a humanoid robot continuously re-planning during each step\citep{Kuffner_2003}. It assumed flat ground and did not provide results on re-planning during each step, but only preliminary data on initial footstep plans. Okada et al. developed a planar segment finder and incorporated that with footstep planning over uneven flat terrain but did not show results for re-planning over multiple steps\citep{Okada_2005}. Cupec et al. used line features in the environment to step over planks and onto uneven flat steps as part of a very reliable 20+ step demo\citep{Cupec_2003}.

Nishiwaki et al. developed a walking behavior that uses realtime perception of the terrain while walking\citep{Nishiwaki_2017}. In this work, the human operator draws a body path using a mixed reality interface. It is able to re-plan footsteps using lidar over rough terrain during walking. This work also supports using a joystick to control the path of the robot. However, the speed and reliability of the results are not presented.

Karkowski et al. presented a realtime perception algorithm that can identify planar and nonplanar regions\citep{Karkowski_2016}. It then demonstrated continuous footstep re-planning to subgoals along the path to the goal. The robot was able to re-plan around obstacles placed in it's way while walking. This system exhibited short pauses in walking during re-planning.

Marion et al. achieve continuous humanoid locomotion using filtered and fused stereo vision\citep{Marion_2016}. They construct a dense point cloud at 10 Hz, segment planar regions in $\sim$615 ms, and plan the next several footsteps in $\sim$445 ms on average. This delay of $\sim$1.3 seconds enables the robot to walk slowly but without pauses over rough terrain.

Lee et al. demonstrate a continuously re-planning walking behavior over flat terrain with gaps\citep{Lee_2021}. Using a depth sensor, they apply RANSAC to segment major planes and put qualifying planar regions into an Octomap for footstep planning. They do this at a high frequency such that they can re-plan footsteps during walking. They show robustness to active terrain disturbance by re-planning a footstep and utilizing the disturbed planar region as a foothold.

\begin{figure*}[!htb]
\centering
    \includegraphics[width=1.6\columnwidth]{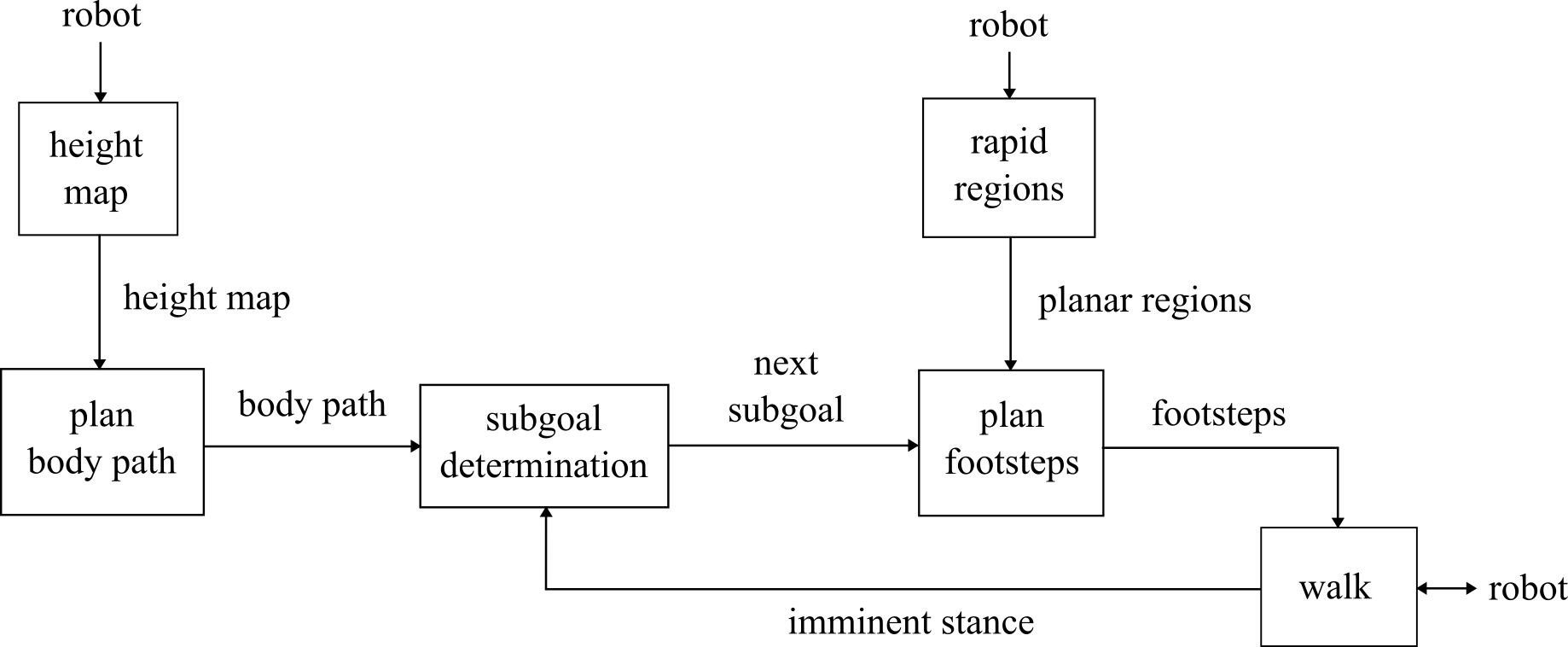}
\caption{The high level control system that we affectionately call ``look-and-step''. This diagram is approximate to give the high level idea behind the system. There are many details that complicate it when considering operator teaming, safety, and edge case handling. We attempt to describe those in the text, but the reader may also consult the freely available source code for the exact implementation, linked to at the end of this paper.}
\vspace{-6mm}
\label{fig:look_and_step_module_control_system_body_path}
\end{figure*}

\section{Look-and-Step Behavior Architecture}

At the highest level, the goal of our look-and-step behavior is simple. 
We want the robot to just start walking, without deliberation, and keep walking without pauses over flat, uneven, and rough terrain. 
To achieve this, the robot must quickly take a look at the immediate terrain in the desired direction, take a step, look again, take a step, and do so repeatedly while walking somewhere.
Using this strategy, the robot can be robust to dynamically changing terrain, uncertainty in far away terrain, and drift in the state estimate.

Though the concept is simple, we found that there are many issues that arise when implementing such a system in a robust way. In the real world, terrain and robot state vary greatly and simple heuristics end up breaking down over a large number of edge cases.
Therefore, we chose to incorporate existing proven implementations of the required components into an over-arching behavior system. However, the assembly of these heavyweight parts presented architectural challenges. Many things needed to happen in parallel and much of the data is time critical. To address these issues formally, we chose to use an event driven architecture inspired by the control systems described in \citep{Brooks_1986}, which features inherent fast response and safe parallelism. In that work, Brooks illustrated control systems as interconnected diagrams of modules with inputs, outputs, suppressors, inhibitors, and resets. We implemented such functionality and built our behavior using this approach.

The control system, illustrated in \autoref{fig:look_and_step_module_control_system_body_path}, is comprised of height mapping, body path planning, sub-goal determination, planar region extraction, footstep planning, and walking. It plans a path for the body once, and then incrementally re-plans footsteps until the robot reaches the goal. The system commands footsteps to the walking controller and coordinates with the robot controller's state.

\begin{figure}[!t]
\centering
    \includegraphics[width=1.0\columnwidth]{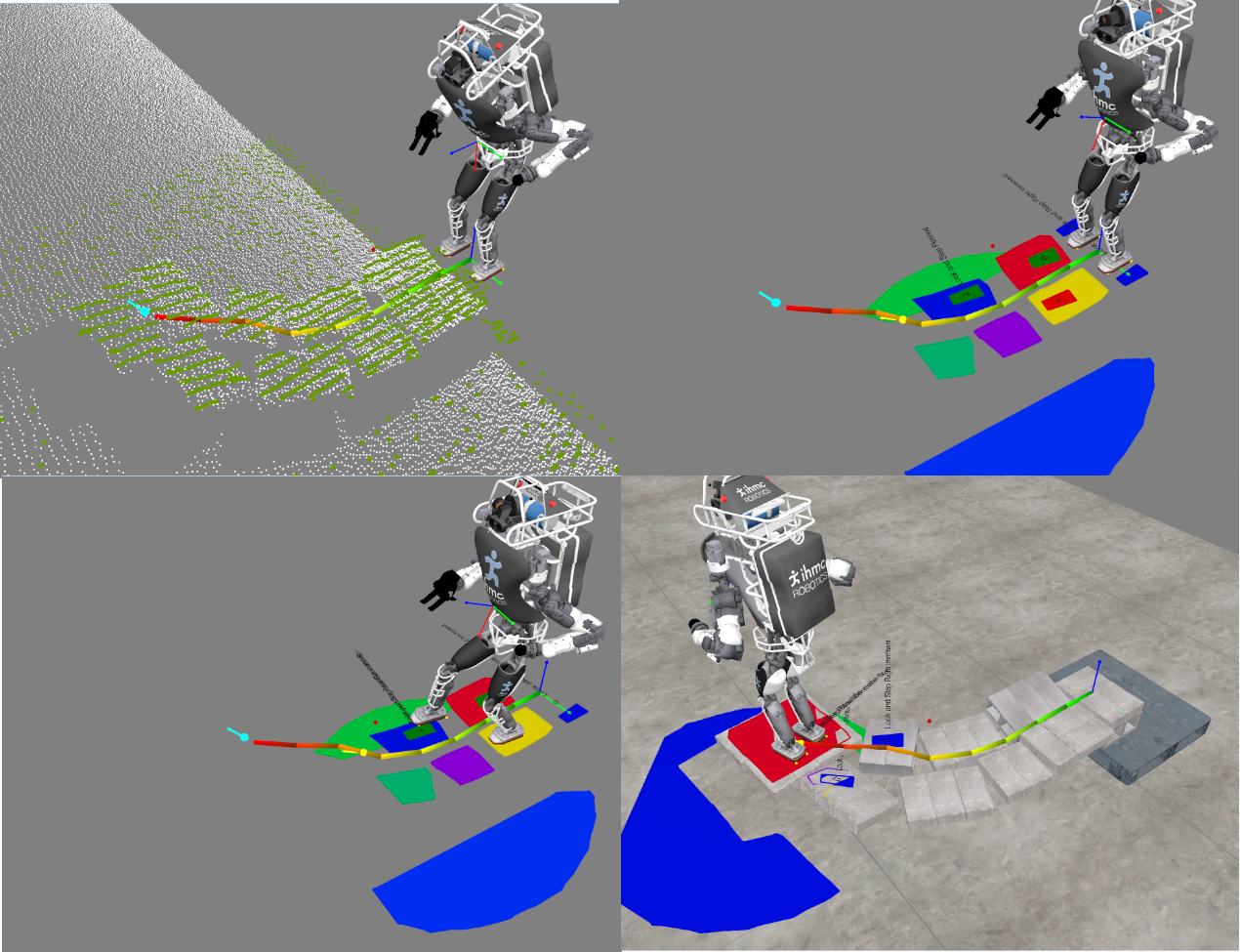}
\caption{Top left: Ouster lidar is used to generate a height map and plan a body path to the human-placed goal. Top right: The initial plan of footsteps to the initial subgoal, placed along the body path. The plan contains 3 steps. Bottom left: The robot is mid-swing. It is currently planning footsteps to the next subgoal around the bend. Bottom right: The robot has squared up it's feet at the end of the behavior, having reached the goal. The ground truth terrain is also shown for reference. Data is from simulation.}
\vspace{-6mm}
\label{fig:look-and-step-stages}
\end{figure}

\subsection{Coactive Design}

We used interdependence design principles to make this behavior observable, predictable, and directable\citep{Johnson_2018}.
Human machine teaming is critical in both the developmental and operational phases. During development, we put the human in the loop as a blocking reviewer so failures can be detected and fixed as early as possible. Later, when allowing continuous re-planning, the operator may still step in to help out.

\begin{figure*}[!t]
\centering
    \includegraphics[width=1.9\columnwidth]{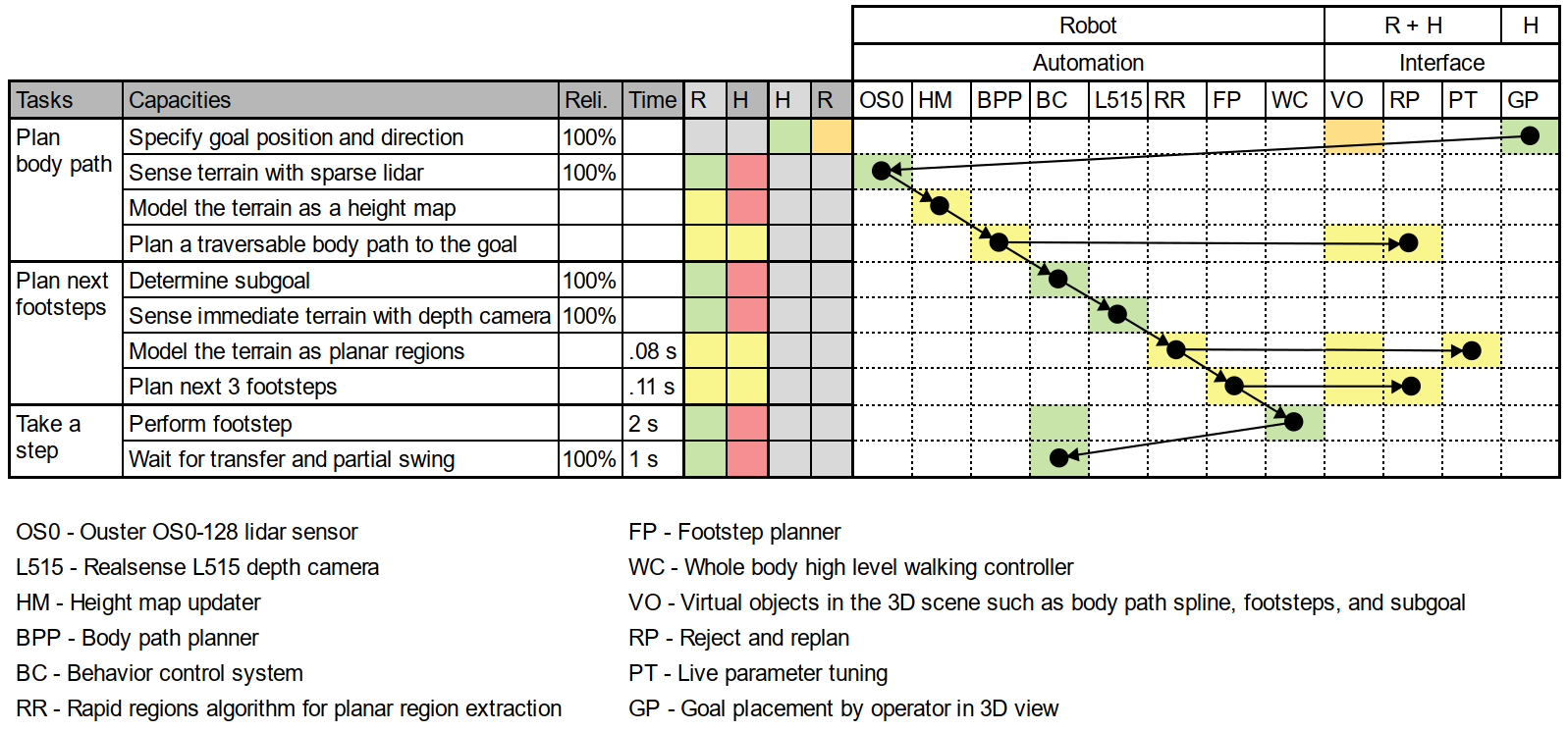}
\caption{Interdependence analysis of the look-and-step behavior. There is a primary autonomous pathway and three supplementary human assisted pathways in which the human can review, tune, reject, and request a re-plan to increase reliability.}
\vspace{-6mm}
\label{fig:interdependence_analysis}
\end{figure*}

As shown in our independence analysis table in \autoref{fig:interdependence_analysis}, the human operator is tasked with providing the goal.
The rest of the plans and actions are calculated by autonomous algorithms.
However, we place a review checkpoint after both the body path and footstep planning tasks.
When those planners complete, the operator is able to preview the results and approve or reject them before they are executed.
If rejected, the planners plan again and display the results for further review. These checkpoints are not only a filter but an opportunity for operator action. The operator can tune parameters during review, providing the ability to work through tough scenarios without aborting the behavior. These features provide the predictability and directablity parts of Coactive Design.

The operator interface displays a live updating 3D scene with the robot, Ouster point cloud, 3D planar regions, and camera views. It also features widgets that show the current state, plot planning statistics, manage the display of virtual objects, and offer buttons and checkboxes for goal placement and planning review, shown in \autoref{fig:look_and_step_ui_widgets}. These features provide the observability part of Coactive Design.

To start the behavior, the operator uses the Ouster lidar point cloud display to place a goal by clicking the ``Place goal'' button and clicking twice in the 3D scene: once for position and once for orientation. To ensure directability, operator review is enforced by default. This is to ensure that autonomous walking only occurs when the operator has explicitly enabled it. To enable the autonomous mode, the operator may disable review and approve the pending plan if there is one, which will cause the behavior to continue autonomously until the goal is reached.

When the robot falls or is otherwise disabled, the behavior detects this event from status messages, resets its internal state, and awaits a new goal from the user. This is required for safety and to prevent the behavior from resuming when the robot is back on it's feet.

\subsection{Body Path Planning over a Height Map}

\begin{figure}[!t]
\centering
    \includegraphics[width=0.9\columnwidth]{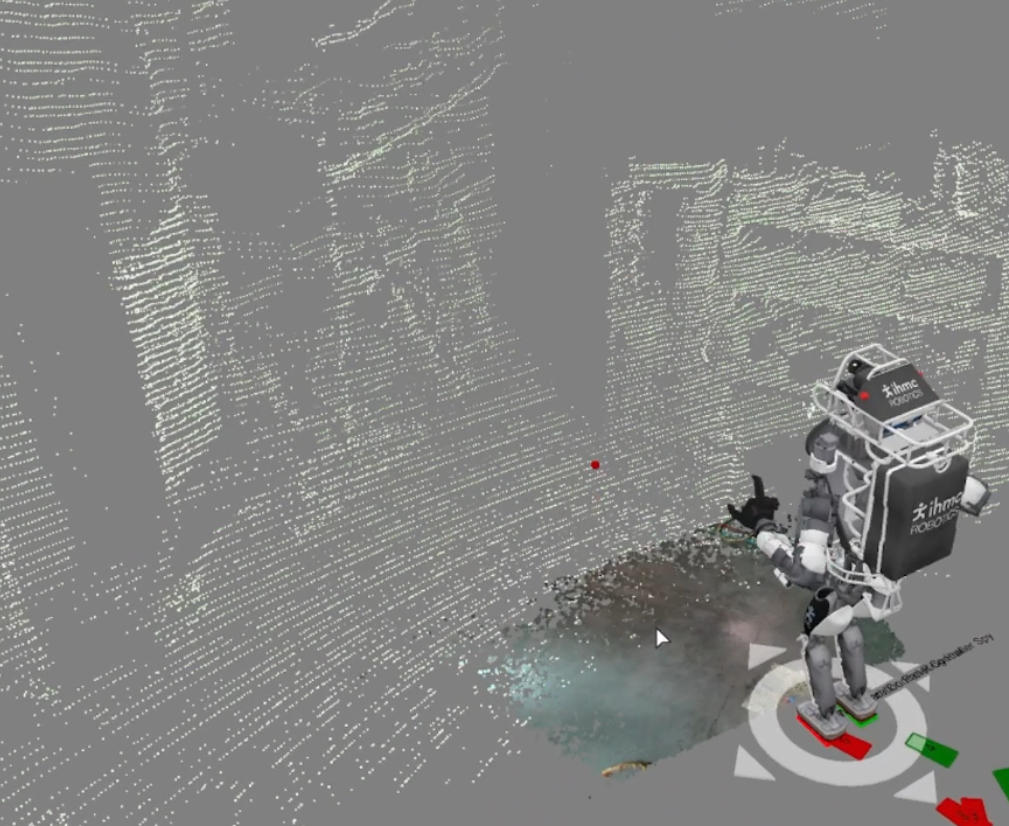}
\caption{The DRC Atlas real robot with Realsense L515 and Ouster OS0-128 data. The two sensors' fields of view overlap in front of the robot. The L515 data is colored and is much denser.}
\vspace{-6mm}
\label{fig:sensor_fovs}
\end{figure}

We use a height map based body path planner developed by McCrory et al.\citep{McCrory_2022}, which first performs an A* graph search with a cost function specifically designed for bipedal locomotion. It then smooths the path by improving the local traversability score further. To generate the height map, we use an Ouster OS0-128 lidar sensor which provides a sparse point cloud with a range of 50 meters, precision of $\pm$ 1.5 at short range to $\pm$ 5.0 cm at long range, and a 10 Hz update rate\citep{ouster}. As seen in \ref{fig:sensor_fovs}, the Ouster's field of view is suitable for long range body path planning, while the Realsense L515 is positioned for planning only in the area immediately in front of the robot. The resulting height map and body path can be seen in \ref{fig:look-and-step-stages}.

\subsection{Subgoal Determination}

We first define the \emph{imminent stance} as the double support stance that the robot will nominally assume when the currently swinging foot lands. In the case that the robot is currently in double support, then the imminent stance is defined as the current stance. This definition is useful to reference the eventual stance conditions which are used for sub-goal determination and footstep planning, as those are done during swing.

Between the re-planning of footsteps, a new sub-goal is determined by projecting the imminent stance's mid-feet pose to the body path and moving it along the path towards the goal by some defined distance. We set this distance to 0.8 meters when on straight stretches and to 0.4 meters when the body path is tightly curving, to more tightly control the resulting footstep plans.
At this point in the behavior, if the robot is near the goal and facing the correct direction, the behavior is reset and does not proceed to footstep planning.

The behavior is also setup to halt here if a significant cluster of Ouster points is detected above the body path, blocking the path. This impassibility detection feature prevents the robot from running into large obstacles and people. The behavior is programmed to resume once the obstacle is gone. This feature can be monitored and disabled in the panel shown in \autoref{fig:look_and_step_ui_widgets}.

\subsection{Perception of Planar Regions}

Perceiving planar regions in a fraction of the foot swing time and during foot swing is critical for continuous walking. In this work we make use of a GPU-accelerated segmentation system for the rapid identification of planar regions presented by Mishra et al.\citep{Mishra_2021}. It is designed to take advantage of depth cameras such as RGB-D structured light and solid state time-of-flight lidar sensors that give full depth images at a high frequency. In this work, we use a Realsense L515 which operates at a resolution of 1024x768 and frame rate of 30 Hz\citep{l515}. The algorithm uses OpenCL kernels to distribute planar region segmentation over many GPU cores, yielding ultra fast update rates. We apply a Gaussian blur, segment the image into a lower resolution patch grid, with normal and Z height information, gathered from the pixel neighborhoods. We then apply a series of depth first searches for regions of normal and height similarity to find locally planar regions from the processed depth image. This is shown visually in \autoref{fig:look_and_step_highlight_composite} and \autoref{fig:look_and_step_simulation_hard_terrain}. These concave regions are decomposed into convex ones and filtered using techniques developed in \citep{Bertrand_2020}. Our implementation runs this process in around 80 ms, as shown in \autoref{tab:part_durations}. On the real robot, we run the process on a Core i5 CPU and NVIDIA 1050 Ti GPU\citep{Mishra_2022}.

The Realsense L515 has field of view of 70$^{\circ}$x55$^{\circ}$, a range of 9 m, and a depth accuracy of 5 mm to 14 mm. We place the sensor on the lower torso, pitched forward by 66$^{\circ}$. By placing the sensor lower, we obtain more accurate depth measurements while sacrificing field of view, highlighted in \ref{fig:sensor_fovs}.



\subsection{Footstep Planning}

\begin{figure}[!t]
\centering
    \includegraphics[width=1.0\columnwidth]{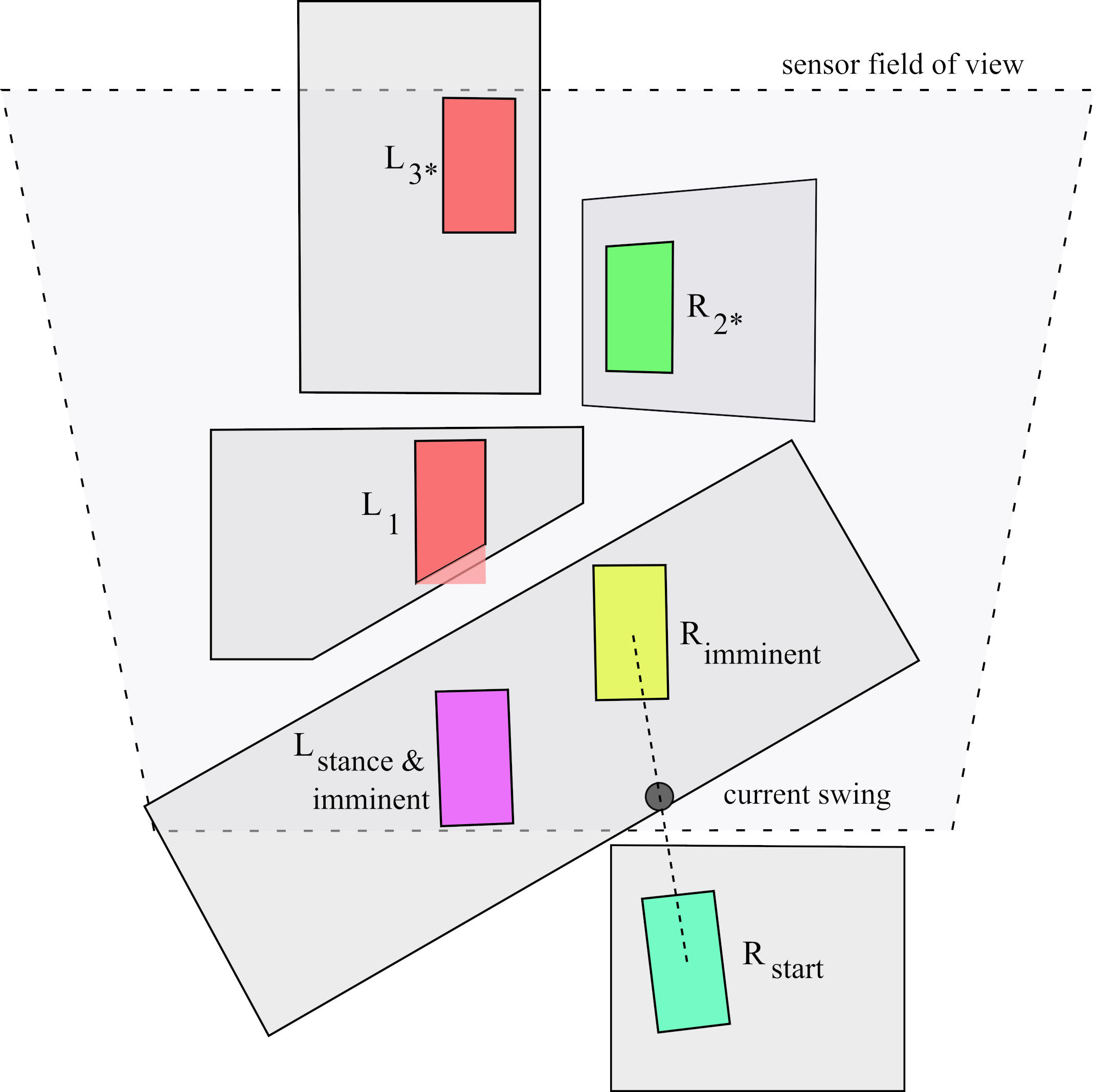}
\caption{This illustration shows a typical planning situation when traversing rough terrain. The footstep planning is invoked while the swing leg is mid stride. The footstep planner is given the ``imminent'' stance feet to start the planning. 3 steps are then planned. Note that steps 2 and 3 are tentative and will be re-planned on the next step.}
\vspace{-6mm}
\label{fig:footstep_planning_with_fov}
\end{figure}

Once an imminent stance, sub-goal, and planar regions are established, we employ a footstep planner capable of optimizing footstep plans over stretches of rough terrain. A typical footstep planning situation is illustrated in \autoref{fig:footstep_planning_with_fov}. Developed by Griffin et al.\citep{Griffin_2019},
the planner performs an A* search over an x-y-$\theta$ grid and can utilize partial footholds to increase capability.
After the initial search, it applies additional local optimizations, such as wiggling footsteps inside planar regions with concave hulls, to maximize walking robustness.
This planner was demonstrated successfully on the NASA Valkyrie and DRC Atlas humanoid robots with plans of 8 to 30 footsteps over distances 3 to 11 meters over rough and irregular terrain.

We task the footstep planner with planning at least 3 steps toward the sub-goal, but also with a timeout. If the timeout is reached then we accept partial progress. If no steps are returned, we consider the planning to have failed. The timeout is set to the expected time remaining in the swing to aim for continuous walking.
During walking, we start the footstep planner halfway through swing with the latest planar regions available at that time.
This takes 200 ms on average which can run in 1/5 of a 1 second swing. If footstep planning fails to plan any steps, it is retried every 2 seconds with a longer timeout. Sometimes tricky situations take longer to plan and this enables a robustness to those.

\subsection{Walking}

Once footsteps have been planned, they are sent to the high level walking controller\citep{Koolen_2016} for execution. Our walking controller is run in a separate realtime process that uses a whole-body momentum-based controller to compute the necessary torques to maintain balance while achieving desired footholds defined in the world frame. 
In our experience, the first three steps most strongly affect the desired dynamic motion, with a drastically reduced effect as this preview window is increased\citep{seyde2018inclusion}. Because of this, we attempt to provide the robot a minimum of two and preferably three steps for every plan. For each subsequent planning update, we replace the steps in the execution queue with the new plan. In this work, we walk with a specified swing time of 1.2 s and a transfer time of 0.8 s.

After tasking the controller with 1 or more steps, the behavior then waits for half of the first step's swing, then triggers the subgoal determination task, closing the loop on continuous perception and walking.

\section{Simulation}

Simulation enables offline development, testing, and evaluation. It is necessary in order to create automated test cases of the system in to warn the developer of regressions. Since simulation can be run quickly and from anywhere, more time can be spent on development.

In order to fully enjoy the benefits of simulation, however, advanced tools are required. We a developed GPU-accelerated depth sensor simulator that is capable of simulating the Ouster OS0-128 and Realsense L515 at full rate and resolution. We also created semi-realistic textured objects that replicate the real objects that we have in the lab. With these tools, we are able to run the same perception and control code as we do on the real robot. This means we can also make improvements to those algorithms offline and test them in varied environments without having to set them up in the real world.

\begin{figure}[!htb]
\centering
    \includegraphics[width=0.9\columnwidth]{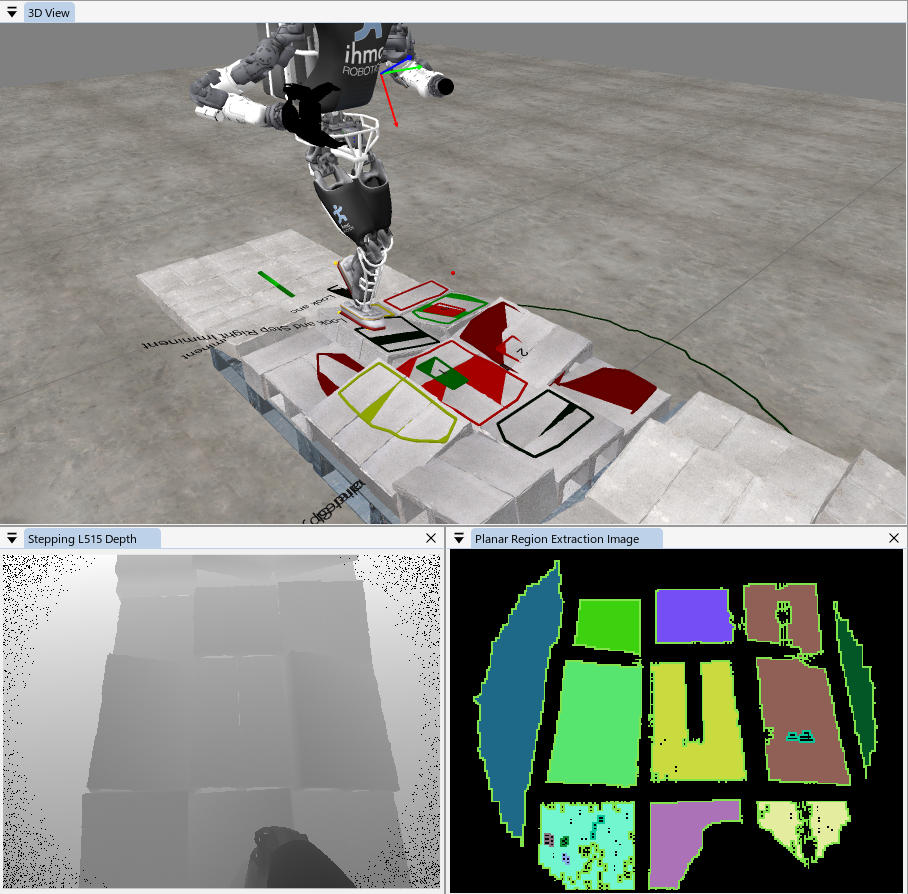}
\caption{Our simulation environment. The robot is shown swinging its left foot in execution of the first in a 3 step plan. The robot's estimated state, footsteps, planar regions, and ground truth terrain are all rendered in 3D. In the lower left, the simulated depth camera's raw depth image is displayed with a noise model applied. In the lower right, the debug planar region extraction image is displayed, showing the state of the algorithm before the regions are converted to 3D.}
\vspace{-6mm}
\label{fig:look_and_step_simulation_hard_terrain}
\end{figure}

Additionally, since physics simulation is computationally expensive and not always able to perform in realtime, we developed a kinematics-only simulator that runs the high level walking controller and integrates the resulting control signals assuming perfect tracking. This yields a simulation that goes through all the nominal motions generated by the controller for given commands. The behavior can then be run at realtime in simulation, in order to exercise the perception and planning algorithms properly and assert that they meet their deadlines. Our simulation environment is shown in \autoref{fig:look_and_step_simulation_hard_terrain}.

\section{Development and Debugging}

Our first line of defense in debugging is to show rich, live visualization. The primary mechanism for this is the 3D view. For planar region extraction, we show a multi-colored debug image shown in the lower right of \autoref{fig:look_and_step_simulation_hard_terrain}. Analyzing this debug image is useful as the first query in a binary search to find issues with that process, as it near the middle of the pipeline. Additionally, parameters can be adjusted via sliders and numerical input widgets and are immediately applied, allowing for a rapid tuning cycle.

\begin{figure}[!hb]
\centering
    \includegraphics[width=0.8\columnwidth]{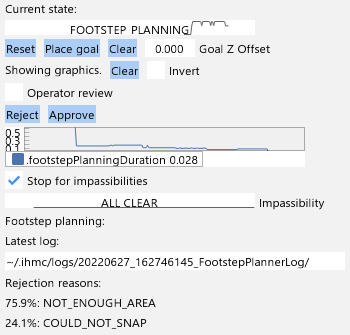}
\caption{The panel of widgets for the look-and-step behavior. It features state plots, buttons, checkboxes, and footstep planner log and failure information.}
\vspace{-2mm}
\label{fig:look_and_step_ui_widgets}
\end{figure}

To debug and tune the body path and footstep planners, we utilize their provided logging system to record every plan to file. We can later open those in a planning log viewer. It has tools to explore the search path of the plan and rerun the plan with different parameters. To more quickly tune the footstep planner, we also compile an abridged summary of the node rejection reasons, such as ``step too high'', ``minimum foothold amount'', or ``too close to cliff'' and print them in the UI, as shown at the bottom of the panel in \autoref{fig:look_and_step_ui_widgets}. Often, this information is enough to inform a parameter adjustment.

Additionally, our high level walking control framework features a sophisticated logging system that logs control variables at 1000 Hz. When the robot loses balance or becomes otherwise unstable, we open the logs in a dedicated visualization program that allows scrubbing through the control variables and balance geometry alongside a synchronized 3rd person log camera video.

We also find graphical debuggers, such as those found in modern integrated development environments, to be critical. These tools can also be used to debug remote processes, which we use often. To shorten the trial and error cycle, we also make use of tools which allow the developer to change code while it is running.

\begin{table}[h]
\caption{Average duration of parts
over a 22 step walk. Results from simulation.}
\centering
\begin{tabular}{c c c} 
 \hline
 Part & Uneven Flat Terrain (s) & Rough Terrain (s) \\
 \hline
 Rapid regions & 0.087 $\pm$ 0.0363  & 0.079 $\pm$ 0.0296  \\ 
 Footstep planning & 0.089 $\pm$ 0.0301 & 0.134 $\pm$ 0.0642 \\
\label{tab:part_durations}
\end{tabular}
\end{table}

\begin{table*}[h]
\caption{Real robot run statistics.}
\centering
\begin{tabular}{c c c c c c c c c}
 \hline
 Total steps & Flat & Uneven flat & Rough & Reactive & Duration (s) & Distance (m) & Avg. speed (m/s) & Avg. step length (m) \\
 \hline
 17 & 8 & 2 & 7 & 0 & 118 & 2.5 & 0.021 & 0.147 \\
 12 & 6 & 6 & 0 & 0 & 30 & 2.1 & 0.070 & 0.175 \\
 12 & 0 & 2 & 10 & 2 & 42 & 3 & 0.071 & 0.250 \\
 23 & 2 & 10 & 11 & 0 & 67 & 5.3 & 0.079 & 0.230 \\
 15 & 0 & 3 & 12 & 0 & 50 & 3.5 & 0.070 & 0.233 \\
 15 & 0 & 7 & 8 & 0 & 42 & 3.5 & 0.083 & 0.233 \\
 18 & 1 & 2 & 15 & 4 & 73 & 3.3 & 0.045 & 0.183 \\
 14 & 2 & 6 & 6 & 0 & 35 & 2.3 & 0.066 & 0.164 \\
 18 & 7 & 11 & 0 & 0 & 42 & 5.5 & 0.131 & 0.306 \\
\label{tab:run_statistics}
\end{tabular}
\end{table*}

\section{Results}

We conducted real world and simulation experiments to evaluate the performance of the system. We did this by constructing cinder block courses with varying levels of difficulty. A combination of flat, uneven flat, and rough terrain was used. We then set the body path goal to the other side of the terrain with operator review disabled. The robot was then left to execute the entire plan autonomously.

\begin{figure*}[!thb]
\centering
    \includegraphics[width=1.5\columnwidth]{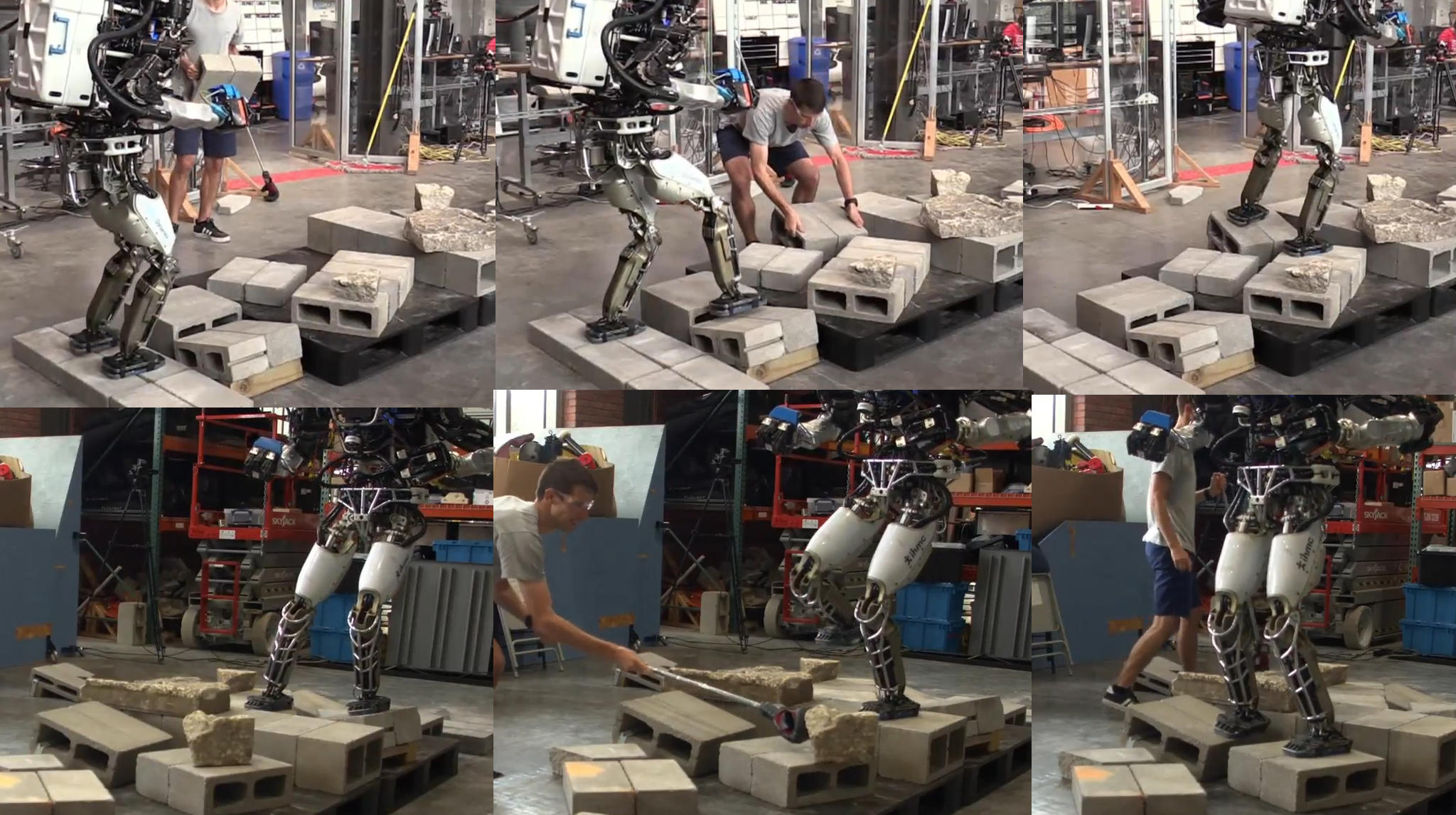}
\caption{This diagram shows a run of the behavior where we added and removed footholds while the robot was walking, just before the robot arrived. The robot was able to utilize the updated environment to achieve forward progress.}
\vspace{-6mm}
\label{fig:adding_and_removing_footholds}
\end{figure*}

In \autoref{tab:run_statistics}, we present statistics on 9 runs of the behavior in autonomous mode on a real DRC Atlas robot. We counted the total number of steps and then categorized those steps by the type of step based on the stance foot, swing start, and swing end footholds for each. Flat steps are defined as where the stance, swing start, and swing end footholds are all coplanar and level. Uneven flat steps are defined as where all three footholds are level but at least one is non-coplanar. Rough steps are defined as where one or more of the footholds are angled and they are not all coplanar. Reactive steps are defined as steps that utilized footholds that were on surfaces that moved while the robot was approaching them. These runs were done over cinder block fields which sometimes contained pieces of cracked concrete, as shown in \autoref{fig:look_and_step_highlight_composite} and \autoref{fig:adding_and_removing_footholds}. All of the body paths were in a straight line and not curving.

We found that the robot walks almost as quickly as predefined footstep plans as presented in \citep{Griffin_2019}. When performing straight line walks, the behavior starts it's first step within 1 second of the operator's initiation. This is a drastic improvement over \citep{Bertrand_2020}, which required a 20 second lidar scan accumulation prior to planning. Additionally, the robot is able to continuously walk farther, in one run walking 5.3 meters over rough terrain without pausing. Our previous capability, presented in \citep{Bertrand_2020}, was only able to facilitate plans of 2 to 3 meters in length. In addition to these improvements, we are now naturally robust to changes in terrain and state estimator drift.

Additionally, we find the planning the 3 next footsteps is feasible during swing for even heavyweight planners, such as our A* planner. We also find that planning more than one step at a time is critical to enable more robust balance by making full use of walking algorithms such as \citep{Koolen_2016}. Though, in order to take advantage of the increased balance, these steps should not change much. We still don't know how to juggle keeping these additional steps consistent while also being able to change them. Reasons for multi-step plan change are both changing terrain and "teetering" results from the footstep planner optimizer.

We also find that depth cameras are suitable for modelling rough terrain while in motion and even while attached to a robot with lots of vibration and shaking at various frequencies. However, we were surprised how much the field of view and mounting angle affected the behavior's capabilities. The sensor must be pointed up and away in order to see far enough for a 3-step plan. However, when mounted in this position it is unable to clearly see it's own feet and the surrounding area. This results in the robot getting stuck if it cannot make progress in the forward direction, with little ability to turn while walking.

At the time of writing, integration of the height map based body path planner is an experimental feature. We were only able to demonstrate it in simulation. Additionally, we were not able to meaningfully measure the planning times because they were varied and dependent on how recently we cleared the height map data.

\section{Discussion and Future Work}

A single distant goal and body path plan is just one way to direct a continuous walking behavior. A useful addition would be a mode for ``carrot on a stick'' guidance, where the robot is tasked with following a continuously updating translation and yaw trajectory, as done in \citep{Nishiwaki_2017}.

The small field of view of the Realsense L515 placed on the lower torso is not sufficient for stepping sideways and backwards and turning in place. Sensors with wider field of views and the incorporation of multiple on-board sensors could help. Sensors with better accuracy could also help if they would allow sensor placement on the head which increases the field of view.

Behavior Trees could potentially replace the event driven architecture presented in this paper and could as serve as a tool to incorporate this behavior into large ones. We think that Behavior Trees might be well suited for constructing large, multi-faceted behaviors via similar abstraction mechanisms as we use in object oriented programming.

Lin and Berenson propose a framework for long horizon planning for humanoid robots in rough terrain environments\citep{Berenson_2021}. They present a multi stage method that plans a torso guiding path and decomposes it into smaller segments with various short range planners. The incorporation of these ideas to continuously re-plan the body path in realtime as we do with footsteps could help to prevent the behavior from getting stuck in local optima.

Kanoulas and Stumpf et al. model terrain as curved contact patches, allowing for more stable footstep planning over rocky terrain\citep{Stumpf_2018}\citep{Kanoulas_2019}. Adding support for this in both the perception and planning modules would greatly enhance the capability of this behavior.

\section{Conclusion}

In this work, we developed a behavior control system that is capable of autonomous, bipedal walking, without pauses for deliberation and planning. We integrated a GPU-accelerated perception algorithm that works for computing planar regions while walking and runs quickly enough to react to moving surfaces. We used a robust A* footstep planner to plan optimal footsteps along a body path. We built the behavior using Coactive Design principles of observability, predictability, and directability, and supported the development with a fully-featured simulation environment. We conducted experiments in simulation and on real robot over flat, uneven, and rough terrain. Our system yields drastic improvements over our previous capabilities, completely removing long pauses, enabling continuous walking without range limits, and enabling reactivity to changing terrain.

\subsection{Acknowledgements}

We would like to thank Perry MacMurray and Allen Reed, who worked on this project during their internships, William Howell and the rest of the team at IHMC Robotics, without whom this work would not have been possible.

\subsection{Source Code and Media}

Our implementation of this walking behavior and the associated modules discussed in this paper can be found on our GitHub at \url{https://github.com/ihmcrobotics}.
The accompanying video can be found at \url{https://youtu.be/-W7wSnhxJIc}.

\bibliography{mybib}

\begin{thebibliography}{10}
\providecommand{\url}[1]{#1}
\csname url@rmstyle\endcsname
\providecommand{\newblock}{\relax}
\providecommand{\bibinfo}[2]{#2}
\providecommand\BIBentrySTDinterwordspacing{\spaceskip=0pt\relax}
\providecommand\BIBentryALTinterwordstretchfactor{4}
\providecommand\BIBentryALTinterwordspacing{\spaceskip=\fontdimen2\font plus
\BIBentryALTinterwordstretchfactor\fontdimen3\font minus
  \fontdimen4\font\relax}
\providecommand\BIBforeignlanguage[2]{{%
\expandafter\ifx\csname l@#1\endcsname\relax
\typeout{** WARNING: IEEEtran.bst: No hyphenation pattern has been}%
\typeout{** loaded for the language `#1'. Using the pattern for}%
\typeout{** the default language instead.}%
\else
\language=\csname l@#1\endcsname
\fi
#2}}

\bibitem{reher2016realizing}
J.~Reher, E.~A. Cousineau, A.~Hereid, C.~M. Hubicki, and A.~D. Ames,
  ``Realizing dynamic and efficient bipedal locomotion on the humanoid robot
  {DURUS},'' in \emph{Robotics and Automation (ICRA), 2016 IEEE International
  Conference on}.\hskip 1em plus 0.5em minus 0.4em\relax IEEE, 2016, pp.
  1794--1801.

\bibitem{Hobart_2020}
C.~G. Hobart, A.~Mazumdar, S.~J. Spencer, M.~Quigley, J.~P. Smith, S.~Bertrand,
  J.~Pratt, M.~Kuehl, and S.~P. Buerger, ``Achieving versatile energy
  efficiency with the wanderer biped robot,'' \emph{IEEE Transactions on
  Robotics}, vol.~36, no.~3, pp. 959--966, 2020.

\bibitem{Gibson_2021}
\BIBentryALTinterwordspacing
G.~Gibson, O.~Dosunmu{-}Ogunbi, Y.~Gong, and J.~W. Grizzle, ``Terrain-aware
  foot placement for bipedal locomotion combining model predictive control,
  virtual constraints, and the {ALIP},'' \emph{CoRR}, 2021. [Online].
  Available: \url{https://arxiv.org/abs/2109.14862}
\BIBentrySTDinterwordspacing

\bibitem{Stumpf_2021}
A.~Stumpf, ``\BIBforeignlanguage{en}{An integrated concept for footstep
  planning and navigation for different types of multi-legged robots in
  challenging environments},'' Ph.D. dissertation, Technische Universit{\"a}t,
  Darmstadt, January 2021.

\bibitem{Fallon_2015}
M.~F. Fallon, P.~Marion, R.~Deits, T.~Whelan, M.~Antone, J.~McDonald, and
  R.~Tedrake, ``Continuous humanoid locomotion over uneven terrain using stereo
  fusion,'' in \emph{2015 IEEE-RAS 15th International Conference on Humanoid
  Robots (Humanoids)}, 2015, pp. 881--888.

\bibitem{Griffin_2019}
\BIBentryALTinterwordspacing
R.~Griffin, G.~Wiedebach, S.~McCrory, S.~Bertrand, I.~Lee, and J.~Pratt,
  ``Footstep planning for autonomous walking over rough terrain.''\hskip 1em
  plus 0.5em minus 0.4em\relax arXiv, 2019. [Online]. Available:
  \url{https://arxiv.org/abs/1907.08673}
\BIBentrySTDinterwordspacing

\bibitem{Beck_2002}
K.~Beck, \emph{Test Driven Development: By Example}.\hskip 1em plus 0.5em minus
  0.4em\relax USA: Addison-Wesley Longman Publishing Co., Inc., 2002.

\bibitem{Ogheneovo_2014}
E.~Ogheneovo, ``On the relationship between software complexity and maintenance
  costs,'' \emph{Journal of Computer and Communications}, vol.~02, pp. 1--16,
  01 2014.

\bibitem{Johnson_2014}
M.~Johnson, J.~M. Bradshaw, P.~J. Feltovich, C.~M. Jonker, M.~B. van Riemsdijk,
  and M.~Sierhuis, ``Coactive design: Designing support for interdependence in
  joint activity,'' \emph{J. Hum.-Robot Interact.}, vol.~3, no.~1, p. 43–69,
  feb 2014.

\bibitem{Mishra_2021}
B.~Mishra, D.~Calvert, S.~Bertrand, S.~McCrory, R.~Griffin, and H.~E. Sevil,
  ``Gpu-accelerated rapid planar region extraction for dynamic behaviors on
  legged robots,'' in \emph{2021 IEEE/RSJ International Conference on
  Intelligent Robots and Systems (IROS)}, 2021, pp. 8493--8499.

\bibitem{McCrory_2022}
\BIBentryALTinterwordspacing
S.~McCrory, B.~Mishra, J.~An, R.~Griffin, J.~Pratt, and H.~E. Sevil, ``Humanoid
  path planning over rough terrain using traversability assessment,'' 2022.
  [Online]. Available: \url{https://arxiv.org/abs/2203.00602}
\BIBentrySTDinterwordspacing

\bibitem{Koolen_2016}
T.~Koolen, S.~Bertrand, G.~Thomas, T.~de~Boer, T.~Wu, J.~Smith, J.~Englsberger,
  and J.~Pratt, ``Design of a momentum-based control framework and application
  to the humanoid robot atlas,'' \emph{International Journal of Humanoid
  Robotics}, vol.~13, no.~01, p. 1650007, 2016.

\bibitem{Stumpf_2014}
A.~Stumpf, S.~Kohlbrecher, D.~C. Conner, and O.~von Stryk, ``Supervised
  footstep planning for humanoid robots in rough terrain tasks using a black
  box walking controller,'' in \emph{2014 IEEE-RAS International Conference on
  Humanoid Robots}, 2014, pp. 287--294.

\bibitem{Atkeson_2018}
C.~G. Atkeson, P.~W.~B. Benzun, N.~Banerjee, D.~Berenson, C.~P. Bove, X.~Cui,
  M.~DeDonato, R.~Du, S.~Feng, P.~Franklin, M.~Gennert, J.~P. Graff, P.~He,
  A.~Jaeger, J.~Kim, K.~Knoedler, L.~Li, C.~Liu, X.~Long, T.~Padir, F.~Polido,
  G.~G. Tighe, and X.~Xinjilefu, \emph{What Happened at the DARPA Robotics
  Challenge Finals}.\hskip 1em plus 0.5em minus 0.4em\relax Cham: Springer
  International Publishing, 2018, pp. 667--684.

\bibitem{Johnson_2017}
M.~Johnson, B.~Shrewsbury, S.~Bertrand, D.~Calvert, T.~Wu, D.~Duran,
  D.~Stephen, N.~Mertins, J.~Carff, W.~Rifenburgh, J.~Smith,
  C.~Schmidt-Wetekam, D.~Faconti, A.~Graber-Tilton, N.~Eyssette, T.~Meier,
  I.~Kalkov, T.~Craig, N.~Payton, S.~McCrory, G.~Wiedebach, B.~Layton,
  P.~Neuhaus, and J.~Pratt, ``Team ihmc's lessons learned from the darpa
  robotics challenge: Finding data in the rubble,'' \emph{Journal of Field
  Robotics}, vol.~34, no.~2, pp. 241--261, 2017.

\bibitem{Kuffner_2003}
J.~Kuffner, S.~Kagami, K.~Nishiwaki, M.~Inaba, and H.~Inoue, ``Online footstep
  planning for humanoid robots,'' in \emph{2003 IEEE International Conference
  on Robotics and Automation (Cat. No.03CH37422)}, vol.~1, 2003, pp. 932--937
  vol.1.

\bibitem{Okada_2005}
K.~Okada, T.~Ogura, A.~Haneda, and M.~Inaba, ``Autonomous 3d walking system for
  a humanoid robot based on visual step recognition and 3d foot step planner,''
  in \emph{Proceedings of the 2005 IEEE International Conference on Robotics
  and Automation}, 2005, pp. 623--628.

\bibitem{Cupec_2003}
R.~Cupec, G.~Schmidt, , and O.~Lorch, ``Vision-guided walking in a structured
  indoor scenario,'' in \emph{Automatika}, vol.~46, 2005, pp. 49--57.

\bibitem{Nishiwaki_2017}
K.~Nishiwaki, J.~Chestnutt, and S.~Kagami, \emph{Autonomous Navigation of a
  Humanoid Robot Over Unknown Rough Terrain}.\hskip 1em plus 0.5em minus
  0.4em\relax Cham: Springer International Publishing, 2017, pp. 619--634.

\bibitem{Karkowski_2016}
P.~Karkowski and M.~Bennewitz, ``Real-time footstep planning using a geometric
  approach,'' in \emph{2016 IEEE International Conference on Robotics and
  Automation (ICRA)}, 2016, pp. 1782--1787.

\bibitem{Marion_2016}
P.~Marion, ``Perception methods for continuous humanoid locomotion over uneven
  terrain,'' Master's thesis, Massachusetts Institute of Technology, 2016.

\bibitem{Lee_2021}
M.~Lee, Y.~Kwon, S.~Lee, J.~Choe, J.~Park, H.~Jeong, Y.~Heo, M.-S. Kim,
  J.~Sungho, S.-E. Yoon, and J.-H. Oh, ``Dynamic humanoid locomotion over rough
  terrain with streamlined perception-control pipeline,'' in \emph{2021
  IEEE/RSJ International Conference on Intelligent Robots and Systems (IROS)},
  2021, pp. 4111--4117.

\bibitem{Brooks_1986}
R.~A. Brooks, ``A robust layered control system for a mobile robot,''
  \emph{IEEE J. Robotics Autom.}, vol.~2, pp. 14--23, 1986.

\bibitem{Johnson_2018}
M.~Johnson, M.~Vignati, and D.~Duran, \emph{Understanding Human-Machine Teaming
  through Interdependence Analysis}.\hskip 1em plus 0.5em minus 0.4em\relax CRC
  Press, 2020, p. Chapter 9.

\bibitem{ouster}
\BIBentryALTinterwordspacing
``Os0 ultra-wide field-of-view lidar sensor for autonomous vehicles and
  robotics.'' [Online]. Available:
  \url{https://ouster.com/products/scanning-lidar/os0-sensor/}
\BIBentrySTDinterwordspacing

\bibitem{l515}
\BIBentryALTinterwordspacing
``Intel realsense lidar camera l515.'' [Online]. Available:
  \url{https://www.intelrealsense.com/lidar-camera-l515/}
\BIBentrySTDinterwordspacing

\bibitem{Bertrand_2020}
S.~Bertrand, I.~Lee, B.~Mishra, D.~Calvert, J.~Pratt, and R.~Griffin,
  ``Detecting usable planar regions for legged robot locomotion,'' in
  \emph{2020 IEEE/RSJ International Conference on Intelligent Robots and
  Systems (IROS)}, 2020, pp. 4736--4742.

\bibitem{Mishra_2022}
B.~Mishra, D.~Calvert, B.~Ortolano, M.~Asselmeier, L.~Fina, S.~McCrory, H.~E.
  Sevil, and R.~Griffin, ``Perception engine using a multi-sensor head to
  enable high-level humanoid robot behaviors,'' in \emph{2022 IEEE/RSJ
  International Conference on Intelligent Robots and Systems (IROS)}, 2022.

\bibitem{seyde2018inclusion}
T.~Seyde, A.~Shrivastava, J.~Englsberger, S.~Bertrand, J.~Pratt, and R.~J.
  Griffin, ``Inclusion of angular momentum during planning for capture point
  based walking,'' in \emph{2018 IEEE International Conference on Robotics and
  Automation (ICRA)}.\hskip 1em plus 0.5em minus 0.4em\relax IEEE, 2018, pp.
  1791--1798.

\bibitem{Berenson_2021}
Y.-C. Lin and D.~Berenson, ``Long-horizon humanoid navigation planning using
  traversability estimates and previous experience,'' in \emph{Autonomous
  Robots}, vol.~45, no.~6, USA, sep 2021, p. 937–956.

\bibitem{Stumpf_2018}
D.~Kanoulas, A.~Stumpf, V.~S. Raghavan, C.~Zhou, A.~Toumpa, O.~Von~Stryk, D.~G.
  Caldwell, and N.~G. Tsagarakis, ``Footstep planning in rough terrain for
  bipedal robots using curved contact patches,'' in \emph{2018 IEEE
  International Conference on Robotics and Automation (ICRA)}, 2018, pp.
  4662--4669.

\bibitem{Kanoulas_2019}
D.~Kanoulas, N.~G. Tsagarakis, and M.~Vona, ``Curved patch mapping and tracking
  for irregular terrain modeling: Application to bipedal robot foot
  placement,'' \emph{Robotics and Autonomous Systems}, vol. 119, pp. 13--30,
  2019.

\end{thebibliography}

\end{document}